\documentclass[letterpaper, 10 pt, conference]{ieeeconf}  

\IEEEoverridecommandlockouts                              

\usepackage{graphics} 
\usepackage{graphicx} 
\usepackage{amsmath} 
\usepackage{amssymb,bm}  
\usepackage[style=ieee]{biblatex}

\DeclareSourcemap{
  \maps{
    \map{
      \pertype{article}
      \step[fieldset=url, null]
      \step[fieldset=doi, null]
      \step[fieldset=issn, null]
      \step[fieldset=isbn, null]
      \step[fieldset=note, null]
      \step[fieldset=editor, null]
      \step[fieldset=urldate, null]
      \step[fieldset=file, null]
    }
  }
}
\DeclareSourcemap{
  \maps{
    \map{
      \pertype{inproceedings}
      \step[fieldset=url, null]
      \step[fieldset=doi, null]
      \step[fieldset=issn, null]
      \step[fieldset=isbn, null]
      \step[fieldset=note, null]
      \step[fieldset=editor, null]
      \step[fieldset=urldate, null]
      \step[fieldset=file, null]
    }
  }
}
\DeclareSourcemap{
  \maps{
    \map{
      \pertype{incollection}
      \step[fieldset=url, null]
      \step[fieldset=doi, null]
      \step[fieldset=issn, null]
      \step[fieldset=isbn, null]
      \step[fieldset=note, null]
      \step[fieldset=editor, null]
      \step[fieldset=urldate, null]
      \step[fieldset=file, null]
    }
  }
}

\title{\LARGE \bf
Dynamic Posture Manipulation During Tumbling\\for Closed-Loop Heading Angle Control 
}

\author{Adarsh Salagame$^{1}$, Eric Sihite$^{2}$, Gunar Schirner$^{1}$, Alireza Ramezani$^{1*}$
\thanks{$^{1}$This author is with the Department of Electrical and Computer Engineering, Northeastern University, Boston MA
		{\tt\small salagame.a, G.Schirner, *a.ramezani@northeastern.edu}}%
  \thanks{$^{2}$This author is with California Institute of Technology, Pasadena CA
		{\tt\small esihite@caltech.edu}}
  \thanks{$*$Indicates corresponding author.}
}

\begin{document}

\maketitle
\thispagestyle{empty}
\pagestyle{empty}

\begin{abstract}
Passive tumbling uses natural forces like gravity for efficient travel. But without an active means of control, passive tumblers must rely entirely on external forces. Northeastern University's COBRA is a snake robot that can morph into a ring, which employs passive tumbling to traverse down slopes. However, due to its articulated joints, it is also capable of dynamically altering its posture to manipulate the dynamics of the tumbling locomotion for active steering. This paper presents a modelling and control strategy based on collocation optimization for real-time steering of COBRA's tumbling locomotion. We validate our approach using Matlab simulations.
\end{abstract}

\section{Introduction}
\label{sec:intro}

Rough terrain locomotion remains a major obstacle for mobile robots, prompting exploration of various design solutions. Of these, legged robots that intermittently interact with their environment to shift the center of mass are the most promising. Unlike wheeled systems with fixed contact points, legged systems can navigate by leveraging their contact-rich locomotion capabilities \cite{park_finite-state_2013,ramezani_atrias_2012}. However, negotiating steep slopes with bumpy surfaces presents distinct challenges that have yet to be fully explored.

Rugged slopes are prevalent on Earth, but perhaps the most scientifically significant examples are found in outer space. The lunar surface, for instance, is riddled with craters that span tens of kilometers in diameter with surface slopes reaching tens of degrees. Their terrain, covered with porous and fluffy regolith (lunar soil), poses formidable locomotion challenges.

\begin{figure}
    \centering
    \includegraphics[width=1.0\linewidth]{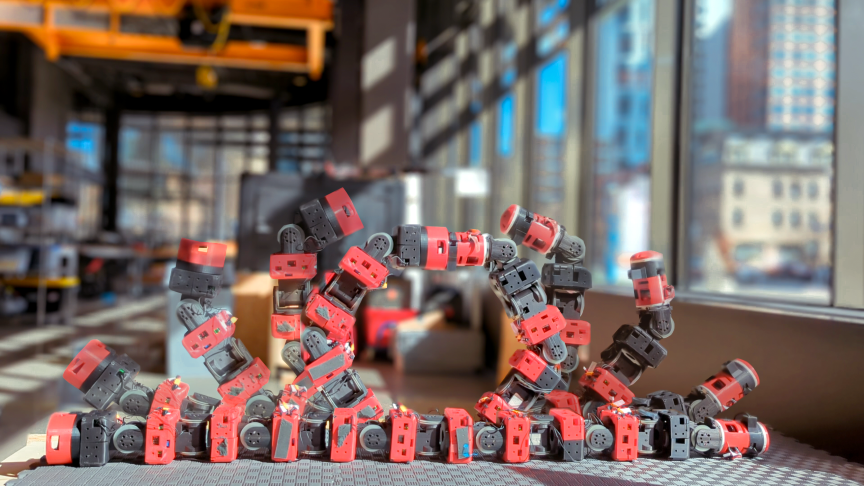}
    \caption{Shows transformation from snake to tumbling configuration in COBRA platform}
    \label{fig:cover-image}
\end{figure}

In 2022, NASA's Innovative Advanced Concepts (NIAC) Program invited US academic institutions to participate in NASA's Breakthrough, Innovative, and Game-changing (BIG) Idea competition \cite{noauthor_big_nodate} by proposing novel mobility systems capable of traversing lunar craters. Northeastern University clinched NASA's top accolade, the Artemis Award, by proposing an articulated robot tumbler named COBRA (Crater Observing Bio-inspired Rolling Articulator) \cite{salagame_how_2023-1} shown in Fig.~\ref{fig:cover-image}. The successful tumbling locomotion demonstrated impressive speed and efficiency over steep slopes in the hills near Pasadena, where the competitions were held. This spurred the formulation of this research question, focusing on active posture manipulation and heading angle steering.

This study specifically aims to elucidate the mathematical model underlying this locomotion challenge and explore control design approaches to address this tracking problem. Before delving into the technical discourse regarding modeling and proposed control design approaches, we briefly survey the literature and previous works pertaining to tumbling robot design.

\subsection{Tumbling Robots}

Tumbling robots have been a subject of interest within the robotic community, with both their merits and limitations well-recognized. Passive tumbling systems, exemplified by the NASA/JPL Mars Tumbleweed Rover \cite{behar_nasajpl_2004}, offer the advantage of minimal energy consumption, making them appealing for remote exploration missions where energy efficiency is paramount. On the other hand, active rolling spherical robots like MIT's Kickbot \cite{batten_kickbot_nodate} boast a low center of gravity and omnidirectional movement capability, rendering them robust to external disturbances and adept at navigating uneven terrain. The ability of spherical robots to roll in any direction also grants them excellent maneuverability in confined spaces.

However, tumbling locomotion presents its own set of challenges. Passive rolling robots often sacrifice controllability in favor of energy efficiency, relying instead on their inherent morphology or posture for maneuvering. Additionally, rolling robots typically utilize their entire body for locomotion, posing difficulties for sensor placement such as cameras and consequently complicating localization and perception tasks.

Early designs, such as Rollo \cite{halme_motion_1996} and the Spherical Mobile Robot (SMR) \cite{reina_rough-terrain_2004}, employed a spherical shell with a diametric spring-mass system to spin a driving wheel, generating movement through a mass imbalance. However, these designs faced reliability issues, particularly in maintaining constant contact between the driving wheel and the sphere. Furthermore, a substantial weight for the central structure was necessary to generate sufficient inertia to propel the system forward.

Subsequent robotic systems, like the University of Pisa's Sphericle \cite{bicchi_introducing_1997} and Festo's Spider Rolling Robot (SRR) \cite{western_golden_2023}, replaced the diametric driving wheel with a car-driven system inside spherical structures. While these designs were promising, they were still unstable from large perturbations on rough terrain.

Another approach to tumbling locomotion involves shifting weights inside a rigid spherical shell to precisely control the 3D position of the center of mass. Examples include the University of Michigan's Spherobot \cite{mukherjee_simple_1999} and the University of Tehran's August Robot \cite{javadi_a_introducing_2004}. However, these systems require additional dead weights to control the center of mass, making them less energy-efficient.

A more energy-efficient method of positioning the center of gravity for tumbling involves using deformable structures. Successful examples include Ritsumeikan University's Deformable Robot \cite{sugiyama_crawling_2006} and Ourobot \cite{paskarbeit_ourobotsensorized_2021} with Articulated Closed Loop. COBRA also adopts the concept of structural deformation to initiate and control tumbling. However, what sets COBRA apart from these examples are its multi-modal locomotion abilities (sidewinding, slithering, and tumbling), field-tested capabilities for fast and dynamic tumbling locomotion on bumpy surfaces, a head-tail locking mechanism to form rugged structures for tumbling, and the ability to actively control posture for steering in two dimensions. This paper primarily focuses on tumbling locomotion and provides an overview of the head-tail locking concept.

\subsection{Overview of COBRA Hardware}

As seen in Fig.~\ref{fig:cover-image}, the COBRA system consists of eleven actuated joints. The front of the robot consists of a head module containing the onboard computing of the system, a radio antenna for communicating with a lunar orbiter, and an inertial measurement unit (IMU) for navigation. At the tail end, there is an interchangeable payload module. The rest of the system consists of identical joint modules each containing a joint actuator and a battery. The robot has a latch mechanism in its head and tail that allows the formation of tumbling structures actively and based on commands tele-communicated to it wirelessly. Due to the large number of body joints, COBRA can actively adjust its posture to regulate its tumbling. This capability was put to the test in open-loop fashion in NASA BIG Idea competitions in 2022.

\begin{figure}
    \centering
    \includegraphics[width=1.0\linewidth]{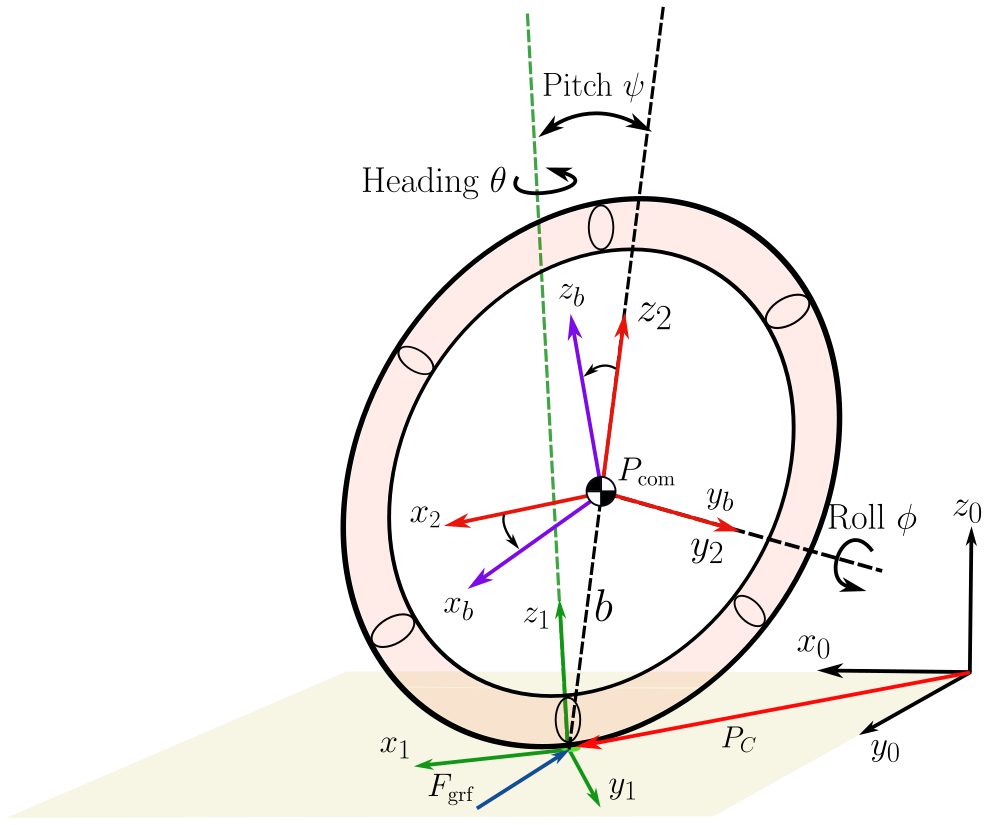}
    \caption{Illustrates model parameters, coordinate frames, etc., used to derive Eqs.~\ref{eq:cascade-model}, \ref{eq:pos-dyn-model}, \ref{eq:M-tbl-dyn}, and \ref{eq:N-tbl-dyn}.}
    \label{fig:rom}
\end{figure}

\section{Reduced-Order Model (ROM) Derivations}
\label{sec:rom}


In the tumbling configuration, COBRA is reduced to an elliptical ring, as depicted in Fig.~\ref{fig:rom}. Throughout these derivations, we make the following assumptions:

\begin{itemize}
    \item The ring has a negligible cross-section area (see Fig.~\ref{fig:rom}).
    \item Mass distribution is uniform.
    \item The shape of the ring is parameterized by the lengths of its principal axes, $u_1$ and $u_2$, controlled by the actuation of joints in the robot (see Fig.~\ref{fig:rom}).
    \item All achieved postures are symmetric, ensuring that the center of the ellipse aligns with the center of mass (CoM).
\end{itemize}

With these assumptions, we determine the inertia tensor of the ring $\mathcal{I}(u_i)$, manipulated by $u_i$. Subsequently, we utilize the inertia tensor to derive the nonholonomic equations of motion of the rolling ring, as the control actions manipulate its posture.

\subsection{Cascade Model}
In our approach to mathematically formulating tumbling servoing using shape manipulations in the ring, we present a cascade nonlinear model of the following form:
\begin{equation}
    \begin{aligned}
    \Sigma_{tbl}&:
    \begin{aligned}
        \dot x &= f(x,y)
    \end{aligned}
    \\
    \Sigma_{pos}&:\left\{
    \begin{aligned}
        \dot \xi &= f_\xi (\xi,u) \\
        y &= h_\xi(\xi)
    \end{aligned}
    \right.
    \end{aligned}
    \label{eq:cascade-model}
\end{equation}
Here, $x$ and $y$ represent the state vector and output function, encapsulating the ring's orientation (Euler angles, CoM position) and mass moment of inertia about its body coordinate axes (x-y-z). The function $f(.)$ governs the dynamics for the states $x$. 

Additionally, $f_\xi$, and $h_\xi$ represent the dynamics governing the internal states $\xi$ (see Section~\ref{sec:rom}). $u$ denotes the actuation effort along the principal axes of the ring as shown in Figs.~\ref{fig:rom} and \ref{fig:deformation}. The cascade model assumes separate governing dynamics for tumbling $\Sigma_{tbl}$ and posture manipulation $\Sigma_{pos}$, considering a specialized case of tumbling that uses internal posture manipulation to maintain the shape of the robot with respect to the inertial frame. In the subsequent subsections, we derive the equations for each model.

\subsection{Governing Dynamics for Posture Manipulation $\Sigma_{pos}$}

As illustrated in Fig.~\ref{fig:rom}, we consider the control actions $u=[u_1,u_2]^\top$ to adjust the length of the principal axes for the ring. Therefore, the objective of this section is to derive the governing equations for posture dynamics which is motivated by posture control from legged robotics \cite{dangol_performance_2020,dangol_feedback_2020,sihite_multi-modal_2023,sihite_optimization-free_2021,dangol_reduced-order-model-based_2021}.

While there is no closed-form equation for the perimeter of an ellipse, several approximations exist that are sufficiently accurate. One such approximation is employed in this work to maintain the perimeter at a fixed value as the length of the axes changes. However, for calculating the inertia tensor for this line mass, the following approach utilizing integrals provides a mathematically simpler method that can subsequently be computed through numerical integration.

\begin{figure}
    \centering
    \includegraphics[width=1.0\linewidth]{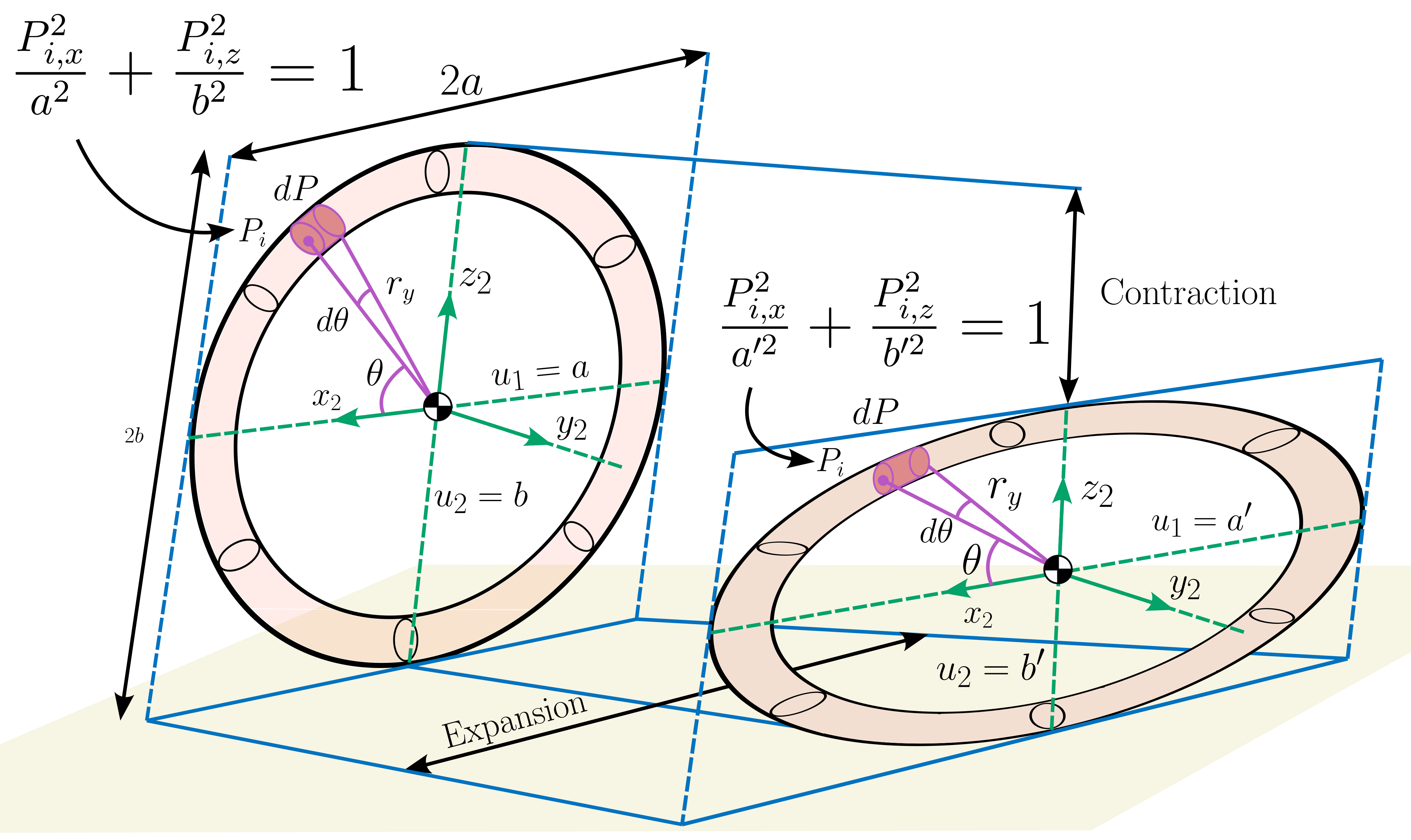}
    \caption{Illustrates posture manipulation by considering two imaginary actuators, denoted as $u_1$ and $u_2$, which act along the principal axes of the ring to induce planar deformations.}
    \label{fig:deformation}
\end{figure}

Consider the general equation of the center line in the ring depicted in Fig.~\ref{fig:deformation} in the x-z plane of the body frame with principal axes of length $a$ and $b$,
\begin{equation}
	\frac{p_{i,x}^2}{a^2} + \frac{p_{i,z}^2}{b^2} = 1
	\label{eq:ellipse_equation}
\end{equation}
where $p_i=[p_{i,x},0,p_{i,z}]^\top$ denotes the body-frame coordinates of a point on the ring. 
Consider the following change of variables:
\begin{equation}
    \begin{aligned}
    & \xi_1=r_y C_\theta \\
    & \xi_2=r_y S_\theta
    \end{aligned}
    \label{eq:change-of-variable}
\end{equation}
where $r_y$ and $\theta$ are polar coordinates and are shown in Fig.~\ref{fig:deformation}. $S_\theta$ and $C_\theta$ denote $\sin \theta$ and $\cos \theta$. We take the time-derivative of the equation above and Eq.~\ref{eq:ellipse_equation}, which yields 
%
\begin{equation}
     \begin{aligned}
    & \dot\xi_1=\dot r_y C_\theta-\xi_2\dot\theta\\
    & \dot\xi_2=\dot r_y S_\theta+\xi_1\dot\theta\\
    & \frac{\xi_1 \dot\xi_1}{a^2} -\frac{2\xi^2_1 u_1}{a^3}+\frac{\xi_2 \dot\xi_2}{b^2}-\frac{2\xi_2^2u_2}{b^3}=0 \\
    &
    \end{aligned}   
    \label{eq:new-ellipse-eq}
\end{equation}
where $\dot a=u_1$ and $\dot b=u_2$. The perimeter of the ring is fixed and given by the following equation
\begin{equation}
        P = \int^{2\pi}_0 \sqrt{a^2 C^2_{\theta} + b^2 S^2_{\theta}}~ d\theta
\end{equation}
therefore, we can write the following relationship between $\dot P$ and $\dot \theta$
\begin{equation}
        \dot P = \left(\sqrt{a^2 C^2_{\theta} + b^2 S^2_{\theta}}\right)\dot \theta = 0
        \label{eq:perimeter-dot}
\end{equation}
This equation constitutes the remaining ordinary differential equations necessary to establish the state-space model for the posture dynamics. By defining $\xi_3=r_y$, $\xi_4=\theta$, $\xi_5=a$, and $\xi_6=b$, and considering Eqs.~\ref{eq:ellipse_equation}, \ref{eq:change-of-variable}, \ref{eq:new-ellipse-eq}, \ref{eq:perimeter-dot}, the state-space model governing the state vector $\xi=\left[\xi_1,\dots,\xi_6\right]^\top$ is given by
\begin{equation}
    \begin{aligned}
        \begin{bmatrix}
        1 & 0 & -C_{\xi_4} & \xi_2 & 0 & 0\\
        1 & 0 & -S_{\xi_4} & \xi_1 & 0 & 0\\
        \frac{\xi_1}{\xi^2_5} & \frac{\xi_2}{\xi^2_6} & 0 & 0 & 0 & 0\\
        0 & 0 & 0 & \gamma(\xi) & 0 & 0\\
        0 & 0 & 0 & 0 & 1 & 0\\
        0 & 0 & 0 & 0 & 0 & 1\\
        \end{bmatrix}
        \begin{bmatrix}
        \dot \xi_1\\
        \dot \xi_2\\
        \dot \xi_3\\
        \dot \xi_4\\
        \dot \xi_5\\
        \dot \xi_6\\
        \end{bmatrix}&=\\
        \begin{bmatrix}
        0 & 0\\
        0 & 0\\
        \frac{2\xi_1^2}{\xi_5^3} & \frac{2\xi_2^2}{\xi_6^3}\\
        0 & 0\\
        1 & 0\\
        0 & 1\\
        \end{bmatrix}
        \begin{bmatrix}
        u_1\\
        u_2
        \end{bmatrix}
    \end{aligned}
    \label{eq:pos-dyn-model}
\end{equation}
where $\gamma(\xi)=\sqrt{\xi_5^2 C^2_{\xi_4} + \xi_6^2 S^2_{\xi_4}}$. The matrix in the left-hand side of Eq.~\ref{eq:pos-dyn-model} is invertible, and therefore, the normal form $\dot \xi=f_\xi(\xi,u)$ can be obtained, which is skipped here. Now, it is possible to show that the mass moments of inertia about the body-frame x, y, and z axes, denoted by $\mathcal{I}_{xx}$, $\mathcal{I}_{yy}$, and $\mathcal{I}_{zz}$, are functions of the hidden state vector $\xi$.

The mass of the differential element on the ring can be calculated assuming uniform distribution as follows:
%
\begin{equation}
	dm = \frac{m}{P} dP = \frac{m}{P} \gamma(\xi) d\xi_4
\end{equation}
where $m$ is the total mass of the elliptical ring. Thus, the mass moment of inertia around each body frame axis can be obtained by:
\begin{equation}
    \begin{aligned}
        \mathcal{I}_{kk} &= \frac{m}{P}\int_{\xi_4} r_k^2\gamma(\xi) d\xi_4,\quad k\in\left\{x, y, z\right\}\\
        r_x&=\xi_3C_{\xi_4}\\
        r_y&=\xi_3\\
        r_z&=\xi_3S_{\xi_4}
    \end{aligned}
\end{equation}
In the equation above, the output function $y=h_\xi(\xi)=[\mathcal{I}_{xx},\mathcal{I}_{yy},\mathcal{I}_{zz}]^\top$ encapsulates the mass moments of inertia. Next, we will derive the equations of motion for the tumbling ring using these posture dynamics as follows.



\subsection{Governing dynamics for Tumbling $\Sigma_{tbl}$}

Consider the ring shown in Fig.~\ref{fig:rom} with imaginary actuators $u_i$ along its two principal axes for posture manipulation. We define the following coordinate frames of reference to describe the motion of the ring:

\begin{itemize}
    \item world frame $x_0$-$y_0$-$z_0$,
    \item contact frame $x_1$-$y_1$-$z_1$ placed at the contact point $p_c$; its z-axis is perpendicular to the ground surface,
    \item gimbal frame $x_2$-$y_2$-$z_2$ placed at the CoM $p_{cm}$; it does not spin with the body,
    \item body frame $x_b$-$y_b$-$z_b$ placed at the CoM $p_{cm}$; it spins with the ring.
\end{itemize}

We incline the contact frame from the world frame at an angle $\alpha$ to create an infinite inclined plane. The orientation $R^0_{b}$ of the ring (body coordinate frame $x_b$-$y_b$-$z_b$) is parameterized using roll, pitch, and yaw angles $\theta$, $\psi$, and $\phi$ expressed as:
\begin{equation}
    R^0_b = R_z(\theta)R_y(\phi)R_x(\psi)
\end{equation}
%
The body frame angular velocity vector $\bm{\omega_b}=[\omega_{b,x},\omega_{b,y},\omega_{b,z}]^\top$ expressed in terms of $\dot \theta$, $\dot \psi$, and $\dot \phi$ is given by:
\begin{equation}
        \begin{aligned}
        &\omega_{b,x}=\dot{\psi} \sin (\theta) \sin (\phi)+\dot{\theta} \cos (\phi)\\
        &\omega_{b,y}=\dot{\psi} \sin (\theta) \cos (\phi)-\dot{\theta} \sin (\phi),\\
        &\omega_{b,z}=\dot{\psi} \cos (\theta)+\dot{\phi}
        \end{aligned}
	\label{eq:omega_B}
\end{equation}
From $\Sigma_{pos}$, the rings principal moment of inertia are given by $y_1$, $y_2$ and $y_3$. The angular momentum of the ring about $p_{cm}$ is given by $\bm{H_b}=y^\top\bm{\omega_b}$. We define the radius of rotation as the vector from $p_{cm}$ to $p_c$. 
Since the ring is pure rolling at the contact point $p_c$, there will be three constraints to consider, including a holonomic $(v_{c,z} = 0)$ and two nonholonomic $(v_{c,x} = 0)$ and $(v_{c,y} = 0)$ constraints, where $\bm{v_c}$ denotes the contact velocity. 
We can formulate the equations of motion by resolving the linear and angular momentum balance concerning the ring's CoM. The resulting equations of motion, derived from applying the balance laws alongside the non-integrable constraints, constitute a set of differential equations describing the ring's orientation and the lateral translation of its CoM over time. We can express this system of equations in the first-order form $\dot x = f(x,y)=M^{-1}(x,y)N(x,y)$ for numerical integration in MATLAB, given by:
\begin{equation}
    M(x,y)=
    \left[\begin{array}{ccc|c}
    y_1 S_\theta & 0 & 0 & \\
    0 & y_2+m r^2 & 0 & \mathbf{0} \\
    \left(y_3+m r^2\right) C_\theta & 0 & y_3+m r^2 & \\
    \hline & \mathbf{0} & & \mathbf{I}
    \end{array}\right]
    \label{eq:M-tbl-dyn}
\end{equation}
where $\mathbf{0}$ and $\mathbf{I}$ are the zero and identity block matrices, respectively. The state vector is $x=[\theta,\psi,\phi,\dot\theta,\dot\psi,\dot\phi,p_{c,x},p_{c,y}]^\top$ and $r$ denotes the distance to the contact point. The nonlinear vector $N$ is given by
\begin{equation}
    N(x,y)=
    \left[\begin{array}{c}
        \left(y_3-2 y_1\right) \dot{\psi} \dot{\theta} C_\theta+y_1 \dot{\theta} \dot{\phi} \\
        \beta_1 \\
        \left(y_3+2 m r^2\right) \dot{\psi} \dot{\theta} S_\theta \\
        \dot{\psi} \\
        \dot{\theta} \\
        \dot{\phi} \\
        \beta_2 \\
        \beta_3
    \end{array}\right]
    \label{eq:N-tbl-dyn}
\end{equation}
where $\beta_i$ are defined in the appendix. This concludes the cascade model derivation. Next, we propose a controller based on the collocation method to utilize $u$ in order to steer the heading angle of the tumbling ring.






\section{Controls}
\label{sec:ctrl}

To solve this controls problem, i.e., the ring's posture dynamics $\Sigma_{pos}$ is recruited to regulate the heading angle in the tumbling dynamics $\Sigma_{tbl}$, we consider the following cost function given by
\begin{equation}
    J = \|\theta-\theta_d\|_2
    \label{eq:cost}
\end{equation}
where $\theta_d$ is the desired heading angle. The cost function $J$ is governed by the cascade model given by Eq.~\ref{eq:cascade-model}.

We perform temporal (i.e., $t_i,~i=1, \ldots, n, \quad 0 \leq t_i \leq t_f$) discretization of Eq.~\ref{eq:cascade-model} to obtain the following system of equations
\begin{equation}
\dot{x}_{i}=f_i(x_i,y_i), \quad i=1, \ldots, n, \quad 0 \leq t_i \leq t_f
\label{eq:disc-model}    
\end{equation}
where $x_i$ embodies the values of the state vector $x$ at i-th discrete time. And, $y_i$ embodies the principal mass moment of inertia of the ring at i-th discrete time. $f_i$ denotes the governing dynamics of tumbling at i-th discrete time.

We stack all of the discrete values from $x_i$ and $y_i$ in the vectors $X = \left[x^\top_1(t_1), \ldots, x_n^\top(t_n)\right]^\top$ and $Y = \left[y^\top_1(t_1), \ldots, y^\top_n(t_n)\right]^\top$.

We consider 2$n$ boundary conditions at the boundaries of $n$ discrete periods to ensure continuity, given by
\begin{equation}
    r_i\left(x(0), x\left(t_f\right), t_f\right)=0, \quad i=1, \ldots, 2 n
\end{equation}
Since we have $3$ entries in $y$, we consider $3$ inequality constraints to ensure the principal mass moment of inertia remains bounded. $g_i$ is given by
\begin{equation}
    g_i(x(t_i), y(t_i), t_i) \geq 0, \quad i=1,2,3, \quad 0 \leq t_i \leq t_f
\end{equation}

\begin{figure*}
    \centering
    \includegraphics[width=0.85\linewidth]{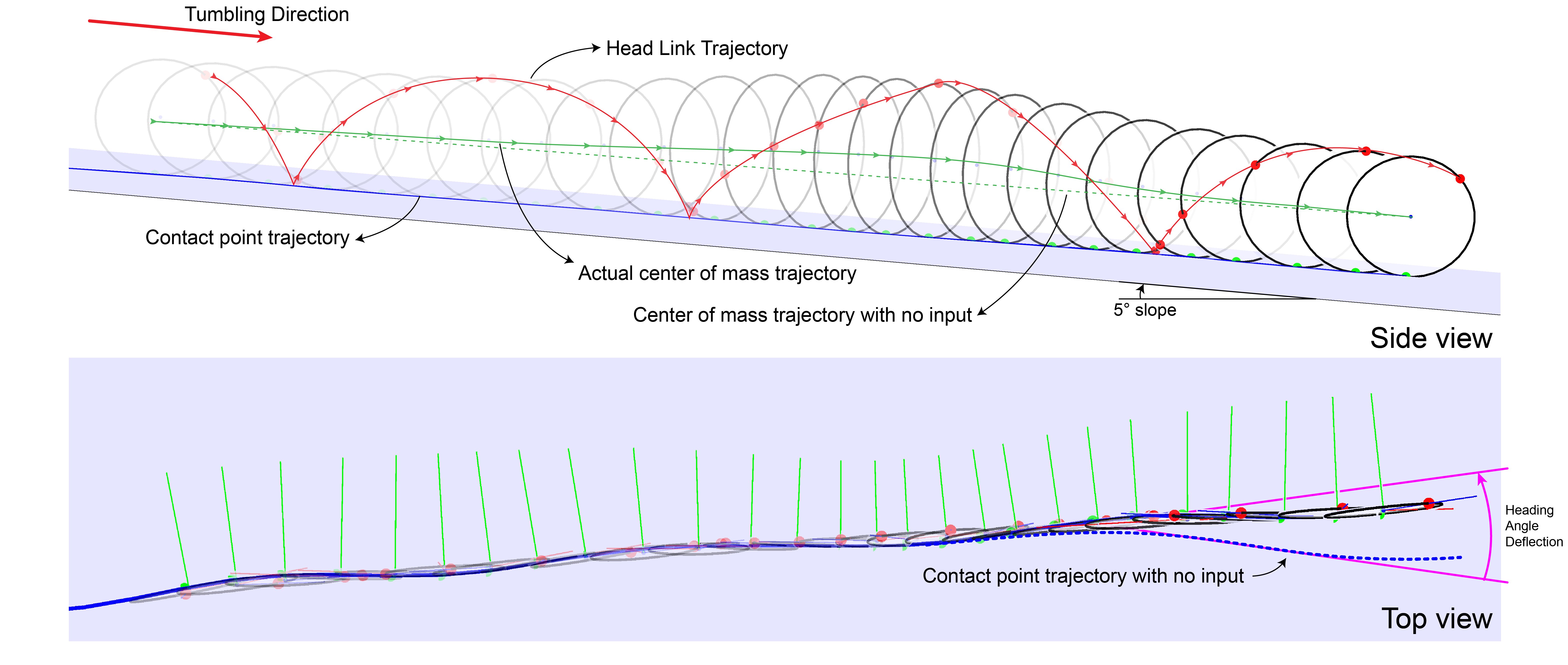}
    \caption{Top figure illustrates the MATLAB simulation of the reduced order model of the robot rolling down a $5^\circ$ incline with control input applied to change the length of the principle axes of the elliptical ring. The bottom figure shows the change in the path of tumbling as an impulse is applied to the axes lengths.}
    \label{fig:sim-snapshots}
\end{figure*}

To approximate the nonlinear dynamics from the tumbling ring, we employ a method based on polynomial interpolations. This method extremely simplifies the computation efforts.  

Consider the $n$ time intervals, as defined previously and given by 
\begin{equation}
0=t_1<t_2<\ldots<t_n=t_f
\end{equation}
We stack the states $x_i$ and principal mass moment of inertia terms $y_i$ from the ring at these discrete times into a single vector denoted by $\mathcal{Y}$ and form a decision parameter vector that the optimizer finds at once. Additionally, we append the final discrete time $t_f$ as the last entry of $\mathcal{Y}$ so that tumbling speed too is determined by the optimizer.
\begin{equation}
\mathcal{Y}=\left[x_1^\top,\dots,x_n^\top,y_1^\top,\dots,y_n^\top,t_f\right]^\top
\end{equation}
We approximate the output function $y_i(t_i)$ at time $t_i \leq t<t_{i+1}$ as the linear interpolation function $\tilde y$ between $y_i(t_i)$ and $y_{i+1}(t_{i+1})$ given by 
\begin{equation}
\tilde y=y_i\left(t_i\right)+\frac{t-t_i}{t_{i+1}-t_i}\left(y_{i+1}\left(t_{i+1}\right)-y_i\left(t_i\right)\right)
\end{equation}
We interpolate the states $x_i(t_i)$ and $x_{i+1}(t_{i+1})$ as well. However, we use a nonlinear cubic interpolation, which is continuously differentiable with $\dot{\tilde x}(s)= f(x(s), y(s), s)$ at $s=t_i$ and $s=t_{i+1}$. 

To obtain $\tilde x$, we formulate the following system of equations:
\begin{equation}
    \begin{aligned}
\tilde x(t) &=\sum_{k=0}^3 c_k^j\left(\frac{t-t_j}{h_j}\right)^k, \quad t_j \leq t<t_{j+1}, \\
c_0^j &=x\left(t_j\right), \\
c_1^j &=h_j f_j, \\
c_2^j &=-3 x\left(t_j\right)-2 h_j f_j+3 x\left(t_{j+1}\right)-h_j f_{j+1}, \\
c_3^j &=2 x\left(t_j\right)+h_j f_j-2 x\left(t_{j+1}\right)+h_j f_{j+1}, \\
\text { where } f_j &:=f\left(x\left(t_j\right), y\left(t_j\right)\right), \quad h_j:=t_{j+1}-t_j .
\end{aligned}
\label{eq:cubic-lobatto}
\end{equation}
The interpolation function $\tilde x$ utilized for $x$ needs to fulfill the continuity at discrete points and at the midpoint of sample times. By examining Eq.~\ref{eq:cubic-lobatto}, it is evident that the derivative terms at the boundaries $t_{i}$ and $t_{i+1}$ are satisfied. Hence, the only remaining constraints in the nonlinear programming problem are the collocation constraints at the midpoint of $t_i-t_{i+1}$ time intervals, the inequality constraints at $t_i$, and the constraints at $t_1$ and $t_f$, all of which are included in the optimization process. We address this optimization problem using MATLAB's fmincon function.

\section{Results and Discussions}
\label{sec:results}

\begin{figure}
    \centering
    \includegraphics[width=0.85\linewidth]{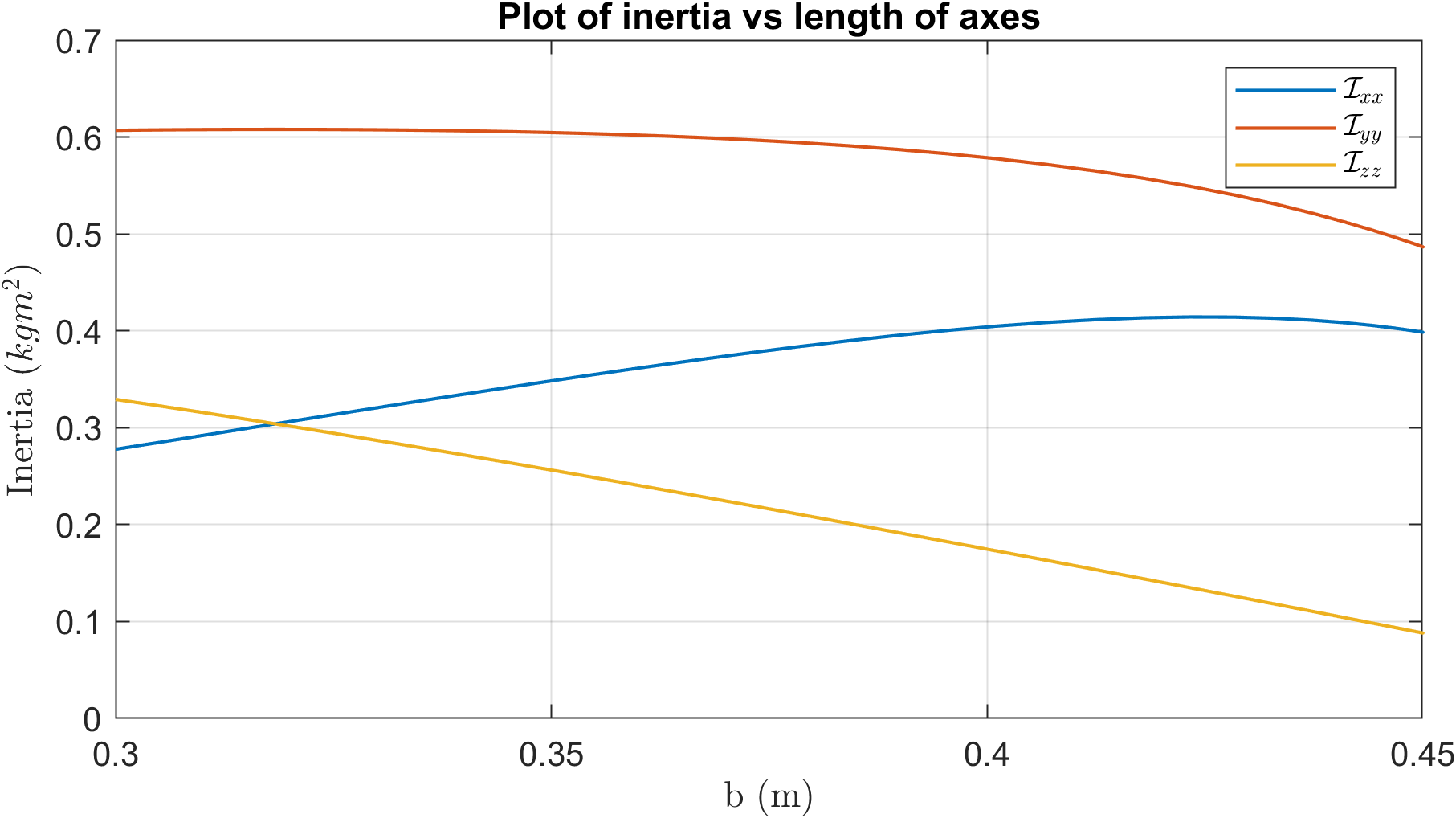}
    \caption{Shows the change in inertia along the principle axes as the lengths of axes are manipulated. }
    \label{fig:inertia-v-b}
\end{figure}

\begin{figure}
    \centering
    \includegraphics[width=0.85\linewidth]{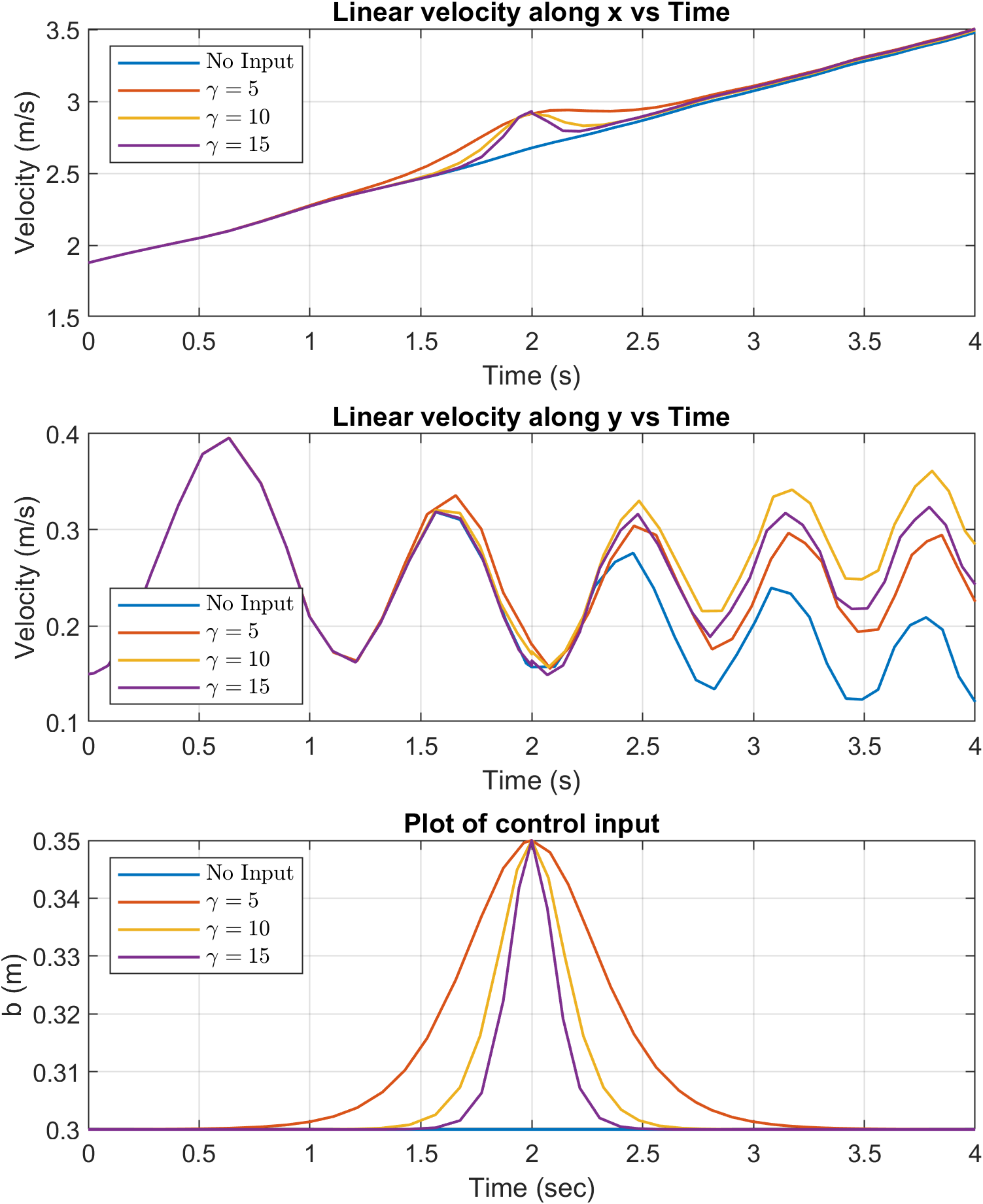}
    \caption{Shows the impact of various impulse inputs on the velocity forward (along $x$) and lateral (along $y$) velocity of the tumbling structure}
    \label{fig:vel}
\end{figure}

\begin{figure}
    \centering
    \includegraphics[width=0.85\linewidth]{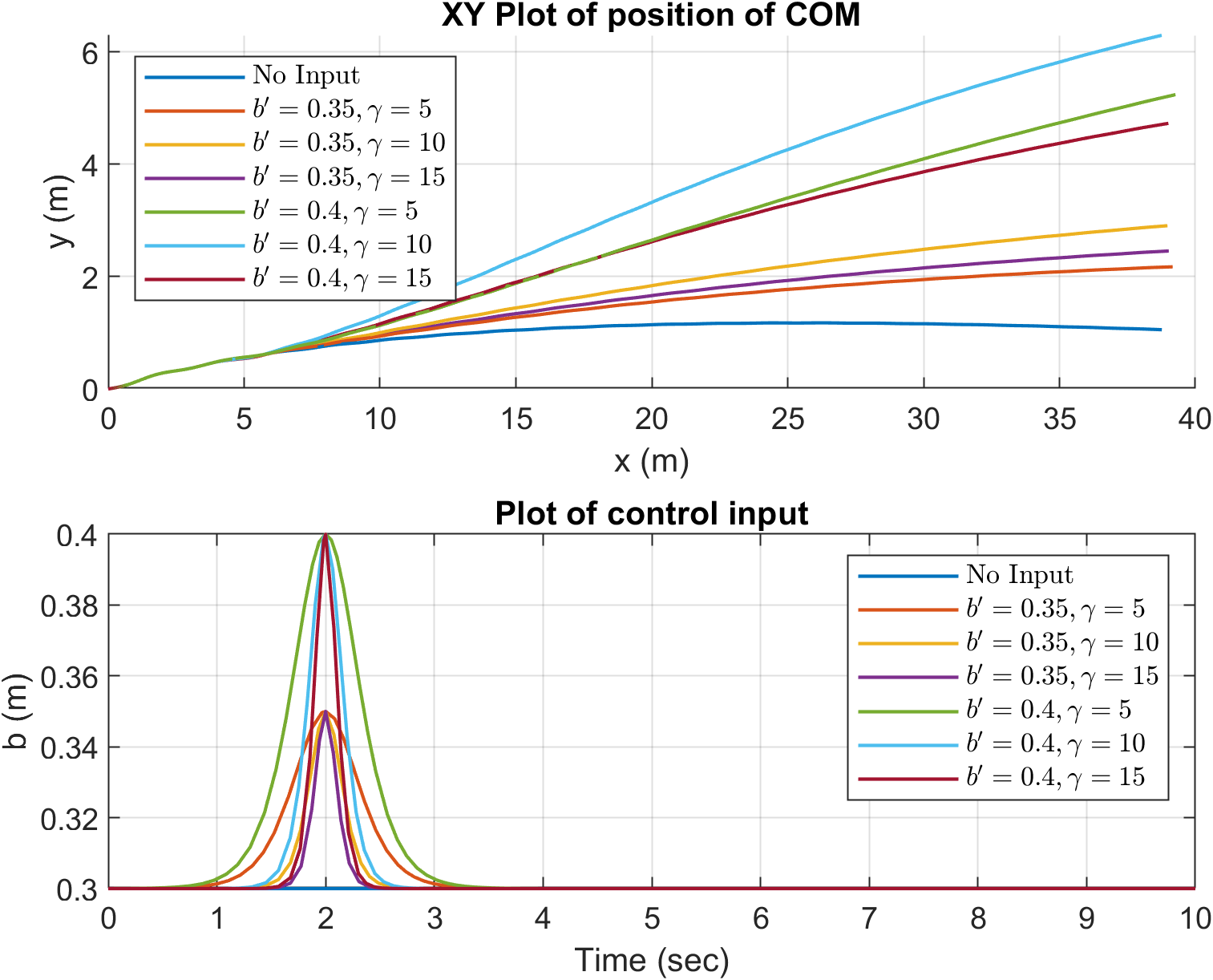}
    \caption{Shows the deflection in the path of the tumbling ellipse from the no control input case upon application of various impulse inputs}
    \label{fig:com-v-bprime}
\end{figure}

We implemented the presented dynamic model in MATLAB and utilized the \textit{ode45} solver to simulate the robot's behavior. The simulation began with the robot positioned on an infinite plane inclined at $5^\circ$ about the inertial $y$ axis with zero heading and pitch angles such that the direction of rolling is along the inertial $x$ axis. Initial rolling and pitching velocities were set to $2\pi$ rad/s and $0.5$ rad/s, respectively. After 2 seconds, an impulse was applied to alter the tumbling trajectory. Figure \ref{fig:sim-snapshots} depicts the simulation setup and showcases snapshots of the robot changing direction upon receiving the input. 

\begin{figure}
    \centering
    \includegraphics[width=0.85\linewidth]{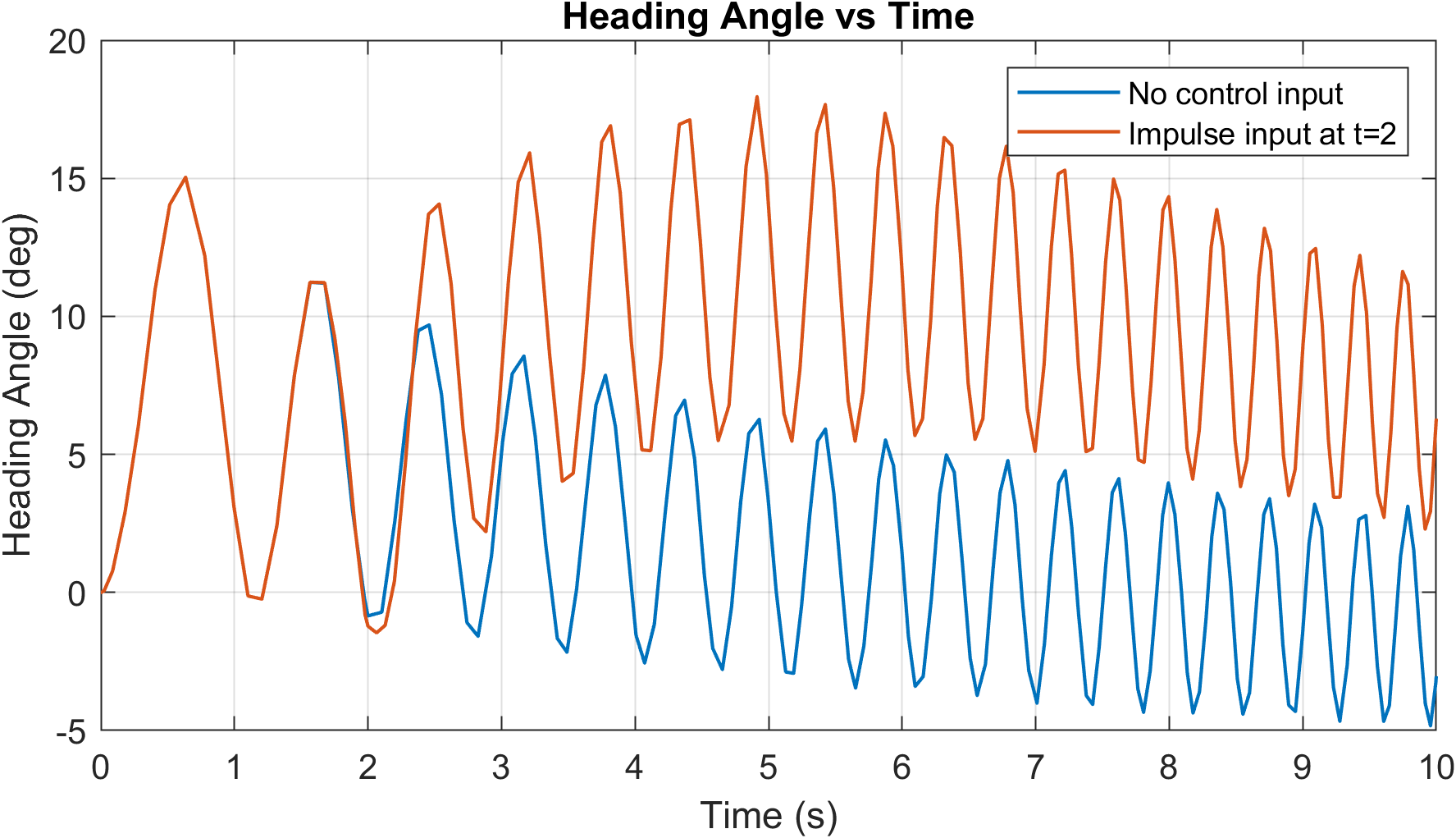}
    \caption{Shows the change in heading angle from the no input case upon application of an impulse input with characteristics: $b^\prime = 0.4~m,~t_0 = 2~s,~\gamma=10$}
    \label{fig:theta}
\end{figure}

The applied input is an impulse signal to the length of the ellipse's vertical principle  axis $b$. The length of the horizontal principle axis $a$ is dynamically calculated based on a fixed perimeter of $2~m$, matching the length of the actual robot. The impulse signal is of the form of the form: 
$$
b(t) = 4b^\prime\sigma(\gamma(t-t_0))(1-\sigma(\gamma(t-t_0))) + b_0
$$ 
where $\sigma$ is the Sigmoid function parameterized by variables $b^\prime$, $t_0$ and $\gamma$ representing the amplitude, time and sharpness of the impulse peak respectively, and $b_0$ refers to the zero input length of the axis $b$. Figure~\ref{fig:inertia-v-b} shows how the inertia of the tumbling structure varies as the axes lengths are changed, which contributes to the change in rolling dynamics during tumbling.  Figure~\ref{fig:vel} shows the response of the tumbling dynamics to various impulse inputs. The linear velocity along $x$ increases and decreases with the input, and lateral velocity along $y$ increases as the heading is changed by the impulse. Figure~\ref{fig:com-v-bprime} shows the $xy$ plot of the center of mass of the robot as it tumbles, with the path changing as impulses of various amplitudes and sharpness are applied. The graph shows sharper impulses with higher amplitudes producing a larger deviation from the nominal trajectory with no control inputs applied. Figure~\ref{fig:theta} shows the corresponding change in heading angle for an input impulse at $t_0 = 2$, with $b^\prime = 0.35$ and $\gamma = 10$.








\section{Concluding Remarks}
\label{sec:conclusion}

This study has established a control framework to steer a simplified model of COBRA during tumbling locomotion by dynamically manipulating its posture. We introduced a cascade model consisting of tumbling and posture dynamics and applied collocation to determine inputs to the cascade model. The simulation results presented herein demonstrate the effectiveness of our approach in adjusting COBRA's heading angle.

Future development will concentrate on integrating this control strategy into a closed-loop system, facilitating real-time steering of the physical robot. Hardware implementation may involve discretizing the simplified model and utilizing COBRA's 11 articulated joints to achieve desired postures. However, this necessitates reliable continuous ground contact estimation to effectively execute posture control. Successful contact estimation and application of the proposed control approach outlined in this paper will enable agile and controllable tumbling locomotion of COBRA in desired directions in real-world scenarios.


\section{Appendix}
\label{sec:app}

The nonlinear terms $\beta_i$ from Eq.~\ref{eq:N-tbl-dyn} are given by 
\[
\begin{aligned}
    \beta_1&=\left(y_1-y_3-m r^2\right) \dot{\psi}^2 S_\theta C_\theta-\left(y_3+m r^2\right) \dot{\psi} \dot{\phi}S_\theta-m g r C_\theta\\
    \beta_2&=-r \dot{\psi} C_\psi C_\theta+r \dot{\theta} S_\psi S_\theta-r \dot{\phi} C_\psi\\ 
    \beta_3&=-r \dot{\psi} S_\psi C_\theta -r \dot{\theta} C_\psi S_\theta -r \dot{\phi} S_\psi
\end{aligned}
\]

\nocite{salagame_how_2023, jiang_hierarchical_2023, salagame_letter_2022, sihite_optimization-free_2021-1, sihite_multi-modal_2023, salagame_quadrupedal_2023, sihite_optimization-free_2021, sihite_unilateral_2021-1, liang_rough-terrain_2021-1, sihite_dynamic_2023}

\printbibliography

\end{document}